\documentclass[conference,compsoc]{IEEEtran}
\ifCLASSOPTIONcompsoc
  \usepackage[nocompress]{cite}
\else
  \usepackage{cite}
\fi

\usepackage{algorithmic}
\usepackage{algorithm}
\usepackage{url}
\usepackage{lineno}
\usepackage{bm}
\usepackage{here}
\usepackage{fancyvrb}
\usepackage{caption}
\usepackage{amsmath}
\usepackage{amsfonts}
\usepackage{ascmac}
\usepackage{comment}
\usepackage{xhfill}
\usepackage{listings}
\makeatletter
\newcommand{\linebreakand}{%
  \end{@IEEEauthorhalign}
  \hfill\mbox{}\par
  \mbox{}\hfill\begin{@IEEEauthorhalign}
}
\makeatother

\begin{document}
\title{Performance Evaluation of General Purpose Large Language Models\\for Basic Linear Algebra Subprograms Code Generation}

\author{
\IEEEauthorblockN{Daichi Mukunoki}
\IEEEauthorblockA{Information Technology Center\\Nagoya University\\
Nagoya, Aichi 466-8601\\
Email: mukunoki@cc.nagoya-u.ac.jp}
\and
\IEEEauthorblockN{Shun-ichiro Hayashi}
\IEEEauthorblockA{Graduate School of Informatics\\Nagoya University\\
Nagoya, Aichi 466-8601\\
Email: hayashi@hpc.itc.nagoya-u.ac.jp}
\and
\IEEEauthorblockN{Tetsuya Hoshino}
\IEEEauthorblockA{Information Technology Center\\Nagoya University\\
Nagoya, Aichi 466-8601\\
Email: hoshino@cc.nagoya-u.ac.jp}
\linebreakand 
\IEEEauthorblockN{Takahiro Katagiri}
\IEEEauthorblockA{Information Technology Center\\Nagoya University\\
Nagoya, Aichi 466-8601\\
Email: katagiri@cc.nagoya-u.ac.jp}
}

\maketitle

\begin{abstract}
Generative AI technology based on Large Language Models (LLM) has been developed and applied to assist or automatically generate program codes. In this paper, we evaluate the capability of existing general LLMs for Basic Linear Algebra Subprograms (BLAS) code generation for CPUs. We use two LLMs provided by OpenAI: GPT-4.1, a Generative Pre-trained Transformer (GPT) model, and o4-mini, one of the o-series of Reasoning models. Both have been released in April 2025. For the routines from level-1 to 3 BLAS, we tried to generate (1) C code without optimization from routine name only, (2) C code with basic performance optimizations (thread parallelization, SIMD vectorization, and cache blocking) from routine name only, and (3) C code with basic performance optimizations based on Fortran reference code. As a result, we found that correct code can be generated in many cases even when only routine name are given. We also confirmed that thread parallelization with OpenMP, SIMD vectorization, and cache blocking can be implemented to some extent, and that the code is faster than the reference code.
\end{abstract}

\IEEEpeerreviewmaketitle

\section{Introduction}
The development of Large Language Models (LLMs) based on Transformers \cite{vaswani2017attention} has led to rapid progress in natural language generative AI technology. For example, OpenAI's ChatGPT\footnote{https://chatgpt.com/}, based on the Generative Pre-trained Transformer (GPT) model \cite{openai2024gpt4technicalreport}, can help solve various problems through natural language interaction via a web interface. This technology can also be applied to program code development, and specialized LLM models have been developed and services based on them have been provided. Given the global shortage of IT human resources, there are high expectations for this kind of AI-based program development support technology. In addition, a paradigm shift is taking place with regard to tasks that should be handled by humans in program development.
\par%%%% 

The goal of code generation with LLM is to automatically generate a desired program from prompt instructions in natural language, etc. In addition to natural language, mathematical expressions for numerical computations, screen images for GUI programs, etc., may be used as input. The ability of LLMs to generate code is highly dependent on the type of target code, in addition to the prompt instructions and information provided. For programming languages that are common and program for common task that have many examples, the generation performance is high due to the large amount of training that has been done.  However, it is insufficient for highly specialized and rarely implemented codes. Numerical codes used in high-performance computing (HPC), which are the target of this study, are one such example. Only a few scientists and engineers develop such numerical codes by themselves. In addition, HPC codes are difficult to implement because they apply many techniques for performance optimization, and only the very best experts are able to implement codes that can achieve performance close to the maximum theoretically possible.
\par%%%

%%%%%%%%%%%%%%%%%%%%%%%%%%%%%%%%%%%%%%%%%%%%%%%%%%%%
\begin{table*}[t]
 \caption{LLM models used in our experiments}
 \label{tab:models}
 \centering
  \begin{tabular}{ll|rr}
   \hline
\multicolumn{2}{l|}{}& GPT-4.1 & o4-mini \\\hline 
Context &Context window & 1,047,576 & 200,000\\
 &Max output tokens & 32,768 & 100,000\\
&Knowledge cutoff & Jun 01, 2024 & Jun 01, 2024\\
Pricing (per 1M tokens)&Input & \$2.00 & \$1.10\\
 &Output & \$8.00 & \$4.40\\
\multicolumn{2}{l|}{Reasoning tokens}&No&Yes\\
   \hline
  \end{tabular}
\end{table*}
%%%%%%%%%%%%%%%%%%%%%%%%%%%%%%%%%%%%%%%%%%%%%%%%%%%%

In this study, we aim to evaluate the capability of existing general LLMs for numerical code generation. This study targets Basic Linear Algebra Subprograms (BLAS) \cite{10.1145/355841.355847}, a de facto standard library that provides basic linear algebra operations. This study uses two existing LLMs: GPT-4.1, the Generative Pre-trained Transformer (GPT) model, as well as the o4-mini model, a reasoning model that produces output based on logical thinking and inference. Both models have been released in April 2025. We generate (1) non-optimized C code from BLAS routine name alone, (2) C code with basic optimizations from BLAS routine name alone, and (3) C code with basic optimizations based on the reference code written in Fortran. The motivation for this study is as follows:
\begin{itemize}
	\setlength{\leftskip}{-15pt} 
	\setlength{\itemsep}{0pt}
	\setlength{\parskip}{0pt}
	\setlength{\itemindent}{0pt}
	\setlength{\labelsep}{8pt}
\item BLAS is a suitable example for the simplest numerical codes.
\item BLAS consists of a large number of routines, ranging in difficulty from vector operations (level-1 BLAS), matrix-vector operations (level-2 BLAS), to matrix-matrix operations (level-3 BLAS).
\item BLAS code is available on the Internet as open source and its specifications are documented in detail. It is possible to consider what LLMs are learning.
\item LLM models used in this study are not coding-specific, but can serve as a baseline for technical research.
\end{itemize} 
\par%%%% 

This paper is organized as follows. Section \ref{sec:related_work} introduces the related work. Section \ref{sec:method} describes the experimental method. Section \ref{sec:result} presents the experimental results and discussion. Finally Section \ref{sec:conclusion} summarizes the paper.
\par%%%% 

\begin{table*}[t]%%%%%%%%%%%%%%%%%%%%%%%%%%%%%%%%%%%%%%
\centering
\caption{Target BLAS routines and supported parameters}
\label{tab:blas}
\begin{tabular}{l|l|l|l|l}
\hline
Level&Routine&Description&Parameters&Equation	\\\hline
1&dasum&sum of magnitudes of vector elements&incx&$v=|x_1|+\cdots +|x_n|$	\\
&daxpy&vector-scalar product and add vector&incx, incy&$y=\alpha x + y$	\\
&ddot&vector-vector dot product&incx, incy&$v=x^Ty$	\\
&idamax&index of element with max absolute value&incx&$v=1^{st}k \ni |\sf{re}(x_k)| + |\sf{im}(x_k)|$	\\
&dnrm2&Euclidean norm of vector&incx&$v = ||x||_2$	\\
&drot&rotation of points in plane&incx, incy&$x_i=cx_i+sy_i$, $y_i=-sx_i+cy_i$	\\
&drotm&modified Givens rotation&incx, incy&$\big[x_i/y_i \big] = H \big[ x_i/y_i \big]$	\\\hline
2&dgemv&general matrix-vector multiply&trans, incx, incy&$y=\alpha Ax+\beta y$	\\
&dger&general matrix rank-1 update&incx, incy&$A=\alpha xy^T+A$	\\
&dsymv&symmetric matrix-vector multiply&uplo, incx, incy&$y=\alpha Ax+\beta y$	\\
&dsyr&symmetric rank-1 update&uplo, incx&$A=\alpha xx^T+A$	\\
&dsyr2&symmetric rank-2 update&uplo, incx, incy&$A=\alpha xy^T+\alpha yx^T+A$	\\
&dtrmv&triangular matrix-vector multiply&uplo, diag, incx&$x=Ax$	\\
&dtrsv&triangular matrix-vector solve&uplo, trans, diag, incx&$x=A^{-1}x$	\\\hline
3&dgemm&general matrix-matrix multiply&transa, transb&$C=\alpha AB+\beta C$	\\
&dsymm&symmetric matrix-matrix multiply&side, uplo&$C+\alpha AB+\beta C$	\\
&dsyrk&symmetric rank-k update&uplo, trans&$C=\alpha  AA^{T}+\beta C$	\\
&dsyr2k&symmetric rank-2k update&uplo, trans&$C=\alpha AB^T+\alpha BA^T+\beta C$	\\
&dtrmm&triangular matrix-matrix multiply&side, uplo, trans, diag&$B=\alpha AB$	\\
&dtrsm&triangular matrix-matrix solve&side, uplo, trans, diag&$B=\alpha A^{-1}B$	\\\hline
\end{tabular}
\end{table*}%%%%%%%%%%%%%%%%%%%%%%%%%%%%%%%%%%%%%%

%%%%%%%%%%%%%%%%%%%%%%%%%%%%%%%%%%%%%%%%%%%%%%%%%%%%
%%%%%%%%%%%%%%%%%%%%%%%%%%%%%%%%%%%%%%%%%%%%%%%%%%%%
%%%%%%%%%%%%%%%%%%%%%%%%%%%%%%%%%%%%%%%%%%%%%%%%%%%%
\section{Related Work}
\label{sec:related_work}
Code generation support by LLMs has already reached the point of practical use. For example, the OpenAI Codex \cite{chen2021evaluatinglargelanguagemodels}, which learns code on GitHub based on the GPT-3 model, is used in the GitHub Copilot\footnote{https://github.com/ features/copilot/}. In addition, a number of models specialized in code generation were proposed, such as CodeGen \cite{nijkamp2023codegenopenlargelanguage} and StarCoder \cite{li2023starcodersourceyou}. At the time of this writing, Anthropic's Claude-4\footnote{https://claude.ai/} claims higher performance than other services in several benchmarks. We can enhance the ability to generate domain-specific code by leveraging Retrieval-Augmented Generation (RAG) as reported in CodeRAG\cite{li2025codeRAGsupportivecoderetrieval}. Code translation from Fortran to C++ \cite{ranasinghe-etal-2025-llm}, and its improvement with RAG \cite{bhattarai2024enhancingcodetranslationlanguage} have been studied. MathCoder\cite{wang2024mathcoder} directly generates codes from a given mathematical problem.
\par %%%

There are several studies aimed to implement HPC codes. For example, the HPC-GPT \cite{10.1145/3624062.3624172}, HPC-Coder \cite{Nichols_2024} (same authors \cite{10.1145/3625549.3658689}), LM4HPC \cite{10.1007/978-3-031-40744-4_2}, and MARCO \cite{rahman2025marcomultiagentoptimizinghpc}. There is also RAG-based ARCS \cite{bhattarai2025arcsagenticretrievalaugmentedcode}), and others. MonoCoder \cite{kadosh2024monocoderdomainspecificcodelanguage} attempts to build a small proprietary model instead of a large-scale LLM.
\par%%%

For matrix computation, QiMeng-GEMM \cite{Zhou_Wen_Chen_Gao_Xiong_Li_Guo_Wu_Chen_2025} demonstrated that it can generate high-performance dense matrix multiplication (GEMM) code for RISC-V CPUs and NVIDIA GPUs. Performance evaluations of the DGEMM and STREAM benchmarks with DeepSeek-R1\cite{deepseekai2025deepseekr1incentivizingreasoningcapability} and GPT-4 have also been performed. ChatBLAS \cite{10820659} uses GPT models to generate level-1 BLAS routines for CPUs (with OpenMP), NVIDIA GPUs (with CUDA), and AMD GPUs (with HIP). The same authors also evaluate the performance of Meta's Llama-2 model and OpenAI GPT-3 in generating code for matrix-vector multiplication and matrix-matrix multiplication \cite{valerolara2023comparingllama2gpt3llms}, Open AI Codex. There is also a case study evaluating the generation of three representative BLAS routines, etc. , by Open AI Codex \cite{10.1145/3605731.3605886}. These are similar to this study as they use prompt engineering methods only, but this study uses a newer LLM model to examine a wide range of routine generation and Fortran code-based generation.
\par %%%

%%%%%%%%%%%%%%%%%%%%%%%%%%%%%%%%%%%%%%%%%%%%%%%%%%%%
%%%%%%%%%%%%%%%%%%%%%%%%%%%%%%%%%%%%%%%%%%%%%%%%%%%%
%%%%%%%%%%%%%%%%%%%%%%%%%%%%%%%%%%%%%%%%%%%%%%%%%%%%
\section{Experimental Method}
\label{sec:method}

This section describes the experimental method for evaluating the capability of LLMs for BLAS code generation. 
\par

%%%%%%%%%%%%%%%%%%%%%%%%%%%%%%%%%%%%%%%%%%%%%%%%%%%% 
\subsection{LLM Models} 
\label{sec:llm}
We use the following two existing LLM models, released in April 2025, provided by OpenAI. Both models are not coding-specific and not the most powerful models but can be a baseline to understand the commonly available technology.
\begin{itemize} 
	\setlength{\leftskip}{-15pt} 
	\setlength{\itemsep}{0pt}
	\setlength{\parskip}{0pt}
	\setlength{\itemindent}{0pt}
	\setlength{\labelsep}{8pt}
 \item \textbf{GPT-4.1} (gpt-4.1-2025-04-14): This is one of the latest GPT models at the time of this writing and is described as a ``GPT model for everyday tasks.'' It prioritizes response time and efficiency over the o-series described next. It supports context lengths of up to 1 million tokens and can handle relatively large codes in code generation.
    \item \textbf{o4-mini} (o4-mini-2025-04-16): Unlike the GPT models, this model is called a reasoning model, which performs internal logical thinking and inference before output, and is described as ``for complex multi-step problems'' compared to the GPT model. It is said to be good at mathematics and complex coding.
\end{itemize} 
Table \ref{tab:models} summarizes both models\footnote{https://platform.openai.com/docs/models/compare}. One token is equivalent to about four letters in English. Reasoning tokens are tokens that are consumed as output tokens in order to perform inference internally before output.
\par

We use the models via Python API. AI agents such as ChatGPT have standard parameters that give some degree of randomness to the output, and return different results each time for a single prompt. We set these parameters to the default settings (temperature=1.0, top\_p=1.0). These parameters cannot be set in o4-mini. In this study, we generate 10 codes for the same prompt. ChatGPT has a memory feature that stores the contents of past conversations, and the output can change accordingly, but this feature does not work when the model is used via an API, as in this study.
\par

%%%%%%%%%%%%%%%%%%%%%%%%%%%%%%%%%%%%%%%%%%%%%%%%%%%% 
\subsection{BLAS Routines} 
We target 20 double-precision (64-bit floating-point operation) routines shown in Table \ref{tab:blas} for the evaluation. BLAS routines can take the following parameters depending on the routine.
\begin{itemize} 
	\setlength{\leftskip}{-15pt} 
	\setlength{\itemsep}{0pt}
	\setlength{\parskip}{0pt}
	\setlength{\itemindent}{0pt}
	\setlength{\labelsep}{8pt}
 \item \textbf{\texttt{incx}}: Increment width of vector X. The vector is stored one by one, incx elements at a time. The same applies to \texttt{incy} for vector Y.
    \item \textbf{\texttt{trans}}: Whether the matrices are transposed or not (`N': as is, `T': transposed, `c': conjugate transposed). The same applies to \texttt{transa} and \texttt{transb} for matrices A and B.
    \item \textbf{\texttt{side}}: Whether matrix A comes to the left of X (`L', e.g. $\sf{op}(A)X = \alpha B$) or to the right (`R', e.g. $X \sf{op}(A) = \alpha B$).
    \item \textbf{\texttt{uplo}}: Whether to use the lower triangle (`L') or the upper triangle (`U') of the matrix.
    \item \textbf{\texttt{diag}}: whether to use a unit-triangular (`U') or not (`N') of the matrix.
\end{itemize} 
Our experiments basically verify the behavior and evaluate the performance for all above parameters. We follow the Fortran BLAS interface; it does not implement the switching between column- and row-major storage of matrices supported by CBLAS, the C language interface to BLAS.
\par %%%

%%%%%%%%%%%%%%%%%%%%%%%%%%%%%%%%%%%%%%%%%%%%%%%%%%%% 
\subsection{Correctness and Performance Evaluation} 
We verify the correctness of the generated code and evaluate the performance using a test program \texttt{run\_tests.py} in BLAS++\cite{blaspp}, a C++ template implementation of BLAS. We embed the generated code into BLAS++. The program considers all possible variations of the parameters of a BLAS routine and exhaustively verifies the behavior. Note that we omit some of the default test items in \texttt{run\_tests.py} in our experiments. For matrix transposition with \texttt{trans}, the conjugate transpose case (i.e., trans=`C') for real routines is not evaluated. For \texttt{incx} and \texttt{incy}, the standard test inputs 1, 2, -1, and -2, respectively, but since negative values are not usually used, we give 1 and 2 alone\footnote{Many generated codes did not work correctly when negative values were given.}. Furthermore, \texttt{run\_tests.py} verifies behavior for changes in problem size. For matrix, non-square patterns (e.g., both $m > n$ and $m < n$ for $m^times n$ matrix) are considered. The number of test cases can be a considerable number for routines with a large number of parameters. For example, \texttt{dgemv} and \texttt{dtrsm} take 128 and 256 test cases, respectively.
\par %%%%

The correctness of the result is evaluated by comparing with the result of the reference BLAS, which is called via CBLAS. For the verification, the problem is initialized with uniform random numbers of (0,1) by the \texttt{larnv} routine in LAPACK. The error is evaluated based on theoretical error analysis and machine epsilon using the relative norm. 
\par %%%%

\texttt{run\_tests.py} also evaluates the performance (in Bytes/s for level-1 and -2 routines and in Flops/s for level-3 routines) based on the execution time. 
\par %%%%

For the performance evaluation, we set a constant problem size which is larger than the cache size: $n=16777216$ for level-1 routines, $m=n=8192$ for level-2 routines, and $m=n=k=2048$ for level-3 routines. \texttt{incx} and \texttt{incy} are evaluated only when both are 1. All codes are compiled with gcc/gfortran 11.3.0 with \texttt{--march=native}, but no optimization options are set. We compare the performance with the reference BLAS implementation included in LAPACK 3.12.1\footnote{https://www.netlib.org/lapack/\#\_lapack\_version\_3\_12\_1\_2}\footnote{Currently reference BLAS implementation is included in the LAPACK package.}, which is not parallelized and optimized..
\par %%%

We evaluate the performance using the Flow Type II supercomputer subsystem at the Information Technology Center, Nagoya University. It has two Intel Xeon Gold 6230 CPUs (20 cores, last level cache 27.5 MB) of the Cascade Lake architecture. The number of OpenMP threads are set to 40, the same as the number of physical cores.
\par %%%

%%%%%%%%%%%%%%%%%%%%%%%%%%%%%%%%%%%%%%%%%%%%%%%%%%%% 
\subsection{Code Generation Prompts} 
Preliminarily stating the experimental results, GPT-4.1 and o4-mini seem to learn the specifications of the BLAS routines in general. However, the generated code is different from the structure of the reference BLAS code written in Fortran. Considering this point, we conduct experiments with the following three cases of prompts. Note that the “\#ROUTINE\#” is the routine name to be generated (e.g., \texttt{dgemm}). The bolded parts of the prompts are the essential instructions, and the rest is instructions for formatting the code for the convenience of incorporating it into BLAS++ for the experiments.

\begin{enumerate}
	\setlength{\leftskip}{-15pt} 
	\setlength{\itemsep}{0pt}
	\setlength{\parskip}{0pt}
	\setlength{\itemindent}{0pt}
	\setlength{\labelsep}{8pt}
    \item Generation of C code without optimization from routine name only (\textbf{NameToCcode}): 
    \begin{itembox}[l]{Prompt}
    ``\textit{\textbf{Implement \#ROUTINE\# routine in BLAS in C language.} Function name must be "GPTBLAS\_\#ROUTINE\#". "GPTBLAS" is capitalized. All function argument names must be in lower case, all must be taken as pointers, and those that have not changed in the code must be marked with const. Insert printf("[gptblas]"); at the beginning of the routine. Do not output anything other than the source code. Do not output Markdown code block symbols either.}''
    \end{itembox}
    \item Generation of C code with basic optimizations from routine name only (\textbf{NameToOptCcode}): 
    \begin{itembox}[l]{Prompt}
    ``\textit{\textbf{Implement \#ROUTINE\# routine in BLAS in C language. Thread parallelization, SIMD vectorization, and cache blocking should be considered for speed-up.} Function name must be "GPTBLAS\_\#ROUTINE\#". "GPTBLAS" is capitalized. All function argument names must be in lower case, all must be taken as pointers, and those that have not changed in the code must be marked with const. Insert printf("[gptblas]"); at the beginning of the routine. Do not output anything other than the source code. Do not output Markdown code block symbols either.}''
    \end{itembox}
    \item Generation of C code with basic optimizations based on the Fortran reference code (\textbf{FrtcodeToOptCcode}): 
    \begin{itembox}[l]{Prompt}
    ``\textit{\textbf{Implement C code that has the same functionality as the attached Fortran code; the specifications are written at the beginning of the Fortran code. Thread parallelization, SIMD vectorization, and cache blocking should be considered for speed-up.} Function name must be "GPTBLAS\_\#ROUTINE\#". "GPTBLAS" is capitalized. All function argument names must be in lower case, all must be taken as pointers, and those that have not changed in the code must be marked with const. If you implement the XERBLA function, it must be named "xerbla" and the function body must not be implemented, but the prototype declaration "void xerbla(const char *srname, const int info);" must be added at the beginning of the code. Use macros named "MIN()" for "min()" and "MAX()" for "max()". Insert printf("[gptblas]"); at the beginning of the routine. Do not output anything other than the source code. Do not output Markdown code block symbols either.}''
    \end{itembox} \par
    In addition, we attach the Fortran reference code that is included in LAPACK 3.12.1. This code also includes the specifications as comments at the beginning of the code. For example, Figure \ref{fig:dsymm.txt} shows a part of the comment of the \texttt{dsymm} routine (it contains some markup tags, but they were left as is). 
\end{enumerate}
\par %%%

Because of the randomness of the output as described in Section \ref{sec:llm}, we generate 10 codes for the same prompt.
\par %%%

\begin{figure}[t]%%%%%%%%%%%%%%%%%%%%%%%%%%%%%%%%%%%%%%
\begin{lstlisting}%[caption=dsymm.f,label=fuga]

*> \brief \b DSYMM
*  =========== DOCUMENTATION ===========
* Online html documentation available at
*            http://www.netlib.org/lapack/explore-html/
*  Definition:
*  ===========
*       SUBROUTINE DSYMM(SIDE,UPLO,M,N,ALPHA,A,LDA,B,LDB,BETA,C,LDC)
*       .. Scalar Arguments ..
*       DOUBLE PRECISION ALPHA,BETA
*       INTEGER LDA,LDB,LDC,M,N
*       CHARACTER SIDE,UPLO
*       ..
*       .. Array Arguments ..
*       DOUBLE PRECISION A(LDA,*),B(LDB,*),C(LDC,*)
*       ..
*> \par Purpose:
*  =============
*> \verbatim
*> DSYMM  performs one of the matrix-matrix operations
*>    C := alpha*A*B + beta*C,
*> or
*>    C := alpha*B*A + beta*C,
*> where alpha and beta are scalars,  A is a symmetric matrix and  B and
*> C are  m by n matrices.
*> \endverbatim
*  Arguments:
*  ==========
*> \param[in] SIDE
*> \verbatim
*>          SIDE is CHARACTER*1
*>           On entry,  SIDE  specifies whether  the  symmetric matrix  A
*>           appears on the  left or right  in the  operation as follows:
*>              SIDE = 'L' or 'l'   C := alpha*A*B + beta*C,
*>              SIDE = 'R' or 'r'   C := alpha*B*A + beta*C,
*> \endverbatim
\end{lstlisting}
\caption{The description of \texttt{dsymm} routine (partially. Authors deleted empty lines).}
\label{fig:dsymm.txt}
\end{figure}%%%%%%%%%%%%%%%%%%%%%%%%%%%%%%%%%%%%%%%%

%%%%%%%%%%%%%%%%%%%%%%%%%%%%%%%%%%%%%%%%%%%%%%%%%%%%
%%%%%%%%%%%%%%%%%%%%%%%%%%%%%%%%%%%%%%%%%%%%%%%%%%%%
%%%%%%%%%%%%%%%%%%%%%%%%%%%%%%%%%%%%%%%%%%%%%%%%%%%%
\section{Result and Discussion}
\label{sec:result}

%%%%%%%%%%%%%%%%%%%%%%%%%%%%%%%%%%%%%%%%%%%%%%%%%%%% 
\subsection{Generation Capability of Correct Code} 
\label{sec:correctness_test}
Table \ref{tab:alltest} shows the number of codes that were able to pass the tests by \texttt{run\_tests.py} out of the 10 generated codes. The failed cases include compilation errors, runtime errors (segmentation fault, etc.), and numerical errors (incorrect result). Overall, o4-mini is superior. The difficulty varies from routine to routine, with complex level-3 routines not succeeding in many cases, but routines with a large number of parameters, such as \texttt{dtrmm} and \texttt{dtrsm}, have many test patterns, which may be one reason why it is more difficult to pass them all.
\par %%%

o4-mini is able to generate C code without optimization from routine name only (NameToCcode) except \texttt{dtrsm}. C code generation with basic optimizations from routine name alone (NameToOptCcode) also succeeded for all level-1 and level-2 routines. GPT-4.1 also succeeds in many cases, although less well than o4-mini. These results indicate that both models have an almost correct understanding of the specifications of the BLAS routines. Interestingly, in many cases, the structure of the generated code differs clearly from the Fortran reference code. For example, for \texttt{dgemm}, the reference code uses \texttt{if} statement to separate four patterns of transposition mode (for \texttt{transa} and \texttt{transb}) combinations, and a triple loop is written for each of them. However, none of the 10 generated codes has such a structure. In many cases, the length of the code is shortened. Furthermore, although the reference code has a code that outputs an error code and terminates when an invalid input (e.g., negative problem size) is written, such code was omitted in most cases except the FrtcodeToOptCcode case, which generated the code with the attached reference Fortran code. Therefore, these results can be inferred that the LLMs do not always learn the reference code as the best correct answer but that they learn the specifications of routines with documents available on the Internet. However, this behavior varies from code to code to some extent. For \texttt{dsyr2k}, generating C code without optimization from routine name only (NameToCcode) by o4-mini generates successfully all 10 codes, but they mostly followed the code structure of the reference Fortran code. 
\par %%%

%%%%%%%%%%%%%%%%%%%%%%%%%%%%%%%%%%%%%%%%%%%%%%%%%%%%
\subsection{Generation Capability of Code with Basic Optimizations}
Tables \ref{tab:perf_level1} (level-1 routines), \ref{tab:perf_level2} (level-2 routines) and \ref{tab:perf_level3} (level-3 routines) show the performance comparison of the generated codes with the reference BLAS codes. The performance is evaluated as GB/s for level-1 and level-2 routines (memory intensive) and GFlops/s for level-3 routines (computationally-intensive). Unlike the evaluation in Section \ref{sec:correctness_test}, this evaluation is conducted on a per-parameter combination basis, which does not necessarily mean that a single code obtained the performance for all parameter cases. Some parameter cases may not work with a single code.
\par %%%

For thread parallelization, OpenMP is used. However, the parallelization was not necessarily applied in the outermost loop. For SIMD vectorization, since the prompt did not specify the target processor, it was unclear what SIMD instructions are available. In most cases, OpenMP's \#pragma omp simd was used, but some codes use the AVX2 and AVX-512 SIMD intrinsics. In most cases, the predefined macros (\_\_AVX\_\_ and \_\_AVX512F\_\_) were used to separate the code for each instruction. Since the CPU used in this experiment was AVX-512 capable, AVX-512 should be used. For cache blocking, a block size of 64 was specified in most cases, but 128 or 256 was specified in some codes. It may be possible to set an appropriate block size by giving the processor specifications at the prompt. 
\par

\begin{table*}[t]%%%%%%%%%%%%%%%%%%%%%%%%%%%%%%%%%%%%%%
\centering
\caption{Result of the correctness test (the number of cases passed with one code out of 10 codes)}
\label{tab:alltest}
\begin{tabular}{l|l|r|r|r|r|r|r}
\hline
&&\multicolumn{2}{c|}{NameToCcode}&\multicolumn{2}{c|}{NameToOptCcode}&\multicolumn{2}{c}{FrtcodeToOptCcode}\\\cline{3-8}
Level&Routine	&GPT-4.1&o4-mini&GPT-4.1&o4-mini&GPT-4.1&o4-mini\\\hline
1&dasum&10&10&9&7&8&8	\\
&daxpy&10&10&10&10&8&9	\\
&ddot&10&10&8&9&6&10	\\
&idamax&3&10&2&9&7&7	\\
&dnrm2&10&10&7&5&8&6	\\
&drot&10&10&8&9&6&9	\\
&drotm&8&5&3&6&8&8	\\\hline
2&dgemv&8&9&3&6&3&4	\\
&dger&10&10&7&7&6&10	\\
&dsymv&8&8 &0&2&0&3	\\
&dsyr&10&8 &0&5&7&7	\\
&dsyr2&8&9 &2&7&6&10	\\
&dtrmv&3&4 &0&5&2&1	\\
&dtrsv&8&9 &0&4&4&7	\\\hline
3&dgemm&10&10 &3&0&5&7	\\
&dsymm&4&8 &3&3&1&4	\\
&dsyrk&3&10 &0&1&4&5	\\
&dsyr2k&3&10 &0&0&7&6	\\
&dtrmm&0&1 &0&0&1&0	\\
&dtrsm&0&0 &0&0&5&1	\\\hline
\end{tabular}
\end{table*}%%%%%%%%%%%%%%%%%%%%%%%%%%%%%%%%%%%%%%
% conference papers do not normally have an appendix

\begin{table*}[t]%%%%%%%%%%%%%%%%%%%%%%%%%%%%%%%%%%%%%%
\centering
\caption{Performance of Level-1 routines (in GB/s, problem size: $n=16777216$). Performance ratio to Ref is shown in parentheses.}
\label{tab:perf_level1}
\begin{tabular}{l|r|rr|rr|rr|rr}
\hline
&&\multicolumn{4}{c|}{NameToOptCcode}&\multicolumn{4}{c}{FrtcodeToOptCcode}\\\cline{3-10}
Routine&Ref&\multicolumn{2}{c|}{GPT-4.1}&\multicolumn{2}{c|}{o4-mini}&\multicolumn{2}{c|}{GPT-4.1}&\multicolumn{2}{c}{o4-mini}	\\\hline
dasum		&6.0&68.0&(11.4x)&63.5&(10.7x)&61.5&(10.3x)&63.6&(10.7x)\\
daxpy		&17.2&77.6&(4.5x)&68.8&(4.0x)&71.9&(4.2x)&67.8&(3.9x)\\
ddot		&11.7&65.8&(5.6x)&63.1&(5.4x)&61.0&(5.2x)&60.8&(5.2x)\\
idamax		&7.1&63.8&(9.0x)&65.1&(9.2x)&60.3&(8.6x)&62.8&(8.9x)\\
dnrm2		&5.1&66.5&(12.9x)&64.0&(12.5x)&60.6&(11.8x)&63.5&(12.4x)\\
drot		&10.6&40.6&(3.8x)&34.1&(3.2x)&34.9&(3.3x)&34.4&(3.3x)\\
drotm		&11.1&39.6&(3.6x)&36.6&(3.3x)&34.7&(3.1x)&34.1&(3.1x)\\\hline
\end{tabular}
\end{table*}%%%%%%%%%%%%%%%%%%%%%%%%%%%%%%%%%%%%%%

\begin{table*}[t]%%%%%%%%%%%%%%%%%%%%%%%%%%%%%%%%%%%%%%
\centering
\caption{Performance of Level-2 routines (in GB/s, problem size: $n=8192$). Performance ratio to Ref is shown in parentheses. Blank indicates that all generated codes have failed.}
\label{tab:perf_level2}
\begin{tabular}{l|l|r|rr|rr|rr|rr}
\hline
&&&\multicolumn{4}{c|}{NameToOptCcode}&\multicolumn{4}{c}{FrtcodeToOptCcode}\\\cline{4-11}
Routine&Parameters & Ref&\multicolumn{2}{c|}{GPT-4.1}&\multicolumn{2}{c|}{o4-mini}&\multicolumn{2}{c|}{GPT-4.1}&\multicolumn{2}{c}{o4-mini}	\\\hline
dgemv&trans=N&9.1&5.4&(0.6x)&30.0&(3.3x)&32.3&(3.5x)&6.6&(0.7x)\\
&trans=T&6.7&68.9&(10.4x)&66.3&(10.0x)&67.6&(10.2x)&65.9&(9.9x)\\
dger&&18.5&46.3&(2.5x)&64.9&(3.5x)&66.4&(3.6x)&64.2&(3.5x)\\
dsymv&uplo=L&6.1&&&4.6&(0.8x)&&&4.3&(0.7x)\\
&uplo=U&6.1&&&5.3&(0.9x)&&&4.7&(0.8x)\\
dsyr&uplo=L&17.4&32.4&(1.9x)&64.0&(3.7x)&64.5&(3.7x)&60.9&(3.5x)\\
&uplo=U&17.5&&&61.5&(3.5x)&64.4&(3.7x)&58.9&(3.4x)\\
dsyr2&uplo=L&7.6&26.4&(3.5x)&63.7&(8.3x)&61.0&(8.0x)&60.2&(7.9x)\\
&uplo=U&15.5&29.0&(1.9x)&59.6&(3.9x)&62.1&(4.0x)&58.6&(3.8x)\\
dtrmv&uplo=L, trans=N, diag=N&7.6&&&7.3&(1.0x)&3.0&(0.4x)&3.0&(0.4x)\\
&uplo=L, trans=N, diag=U&7.5&&&7.7&(1.0x)&2.9&(0.4x)&2.9&(0.4x)\\
&uplo=L, trans=T, diag=N&6.5&&&56.9&(8.8x)&3.3&(0.5x)&3.2&(0.5x)\\
&uplo=L, trans=T, diag=U&6.5&&&56.3&(8.7x)&3.3&(0.5x)&3.2&(0.5x)\\
&uplo=U, trans=N, diag=N&9.0&&&7.6&(0.8x)&3.1&(0.3x)&3.1&(0.3x)\\
&uplo=U, trans=N, diag=U&8.8&&&7.6&(0.9x)&3.1&(0.3x)&3.1&(0.4x)\\
&uplo=U, trans=T, diag=N&6.1&&&56.7&(9.3x)&3.2&(0.5x)&3.2&(0.5x)\\
&uplo=U, trans=T, diag=U&6.1&&&56.4&(9.3x)&3.2&(0.5x)&3.2&(0.5x)\\
dtrsv&uplo=L, trans=N, diag=N&8.4&12.0&(1.4x)&17.5&(2.1x)&4.2&(0.5x)&3.5&(0.4x)\\
&uplo=L, trans=N, diag=U&8.5&12.2&(1.4x)&15.4&(1.8x)&4.4&(0.5x)&3.4&(0.4x)\\
&uplo=L, trans=T, diag=N&6.2&&&26.6&(4.3x)&3.1&(0.5x)&3.1&(0.5x)\\
&uplo=L, trans=T, diag=U&6.2&&&26.8&(4.3x)&3.1&(0.5x)&3.1&(0.5x)\\
&uplo=U, trans=N, diag=N&7.7&11.7&(1.5x)&15.1&(2.0x)&4.3&(0.6x)&3.0&(0.4x)\\
&uplo=U, trans=N, diag=U&7.8&12.4&(1.6x)&15.1&(1.9x)&4.4&(0.6x)&3.1&(0.4x)\\
&uplo=U, trans=T, diag=N&6.5&&&27.5&(4.2x)&3.3&(0.5x)&3.3&(0.5x)\\
&uplo=U, trans=T, diag=U&6.5&&&27.1&(4.1x)&3.2&(0.5x)&3.3&(0.5x)\\\hline
\end{tabular}
\end{table*}%%%%%%%%%%%%%%%%%%%%%%%%%%%%%%%%%%%%%%

\begin{table*}[t]%%%%%%%%%%%%%%%%%%%%%%%%%%%%%%%%%%%%%%
\centering
\caption{Performance of Level-3 routines (in GFlops/s, problem size: $n=2048$). Performance ratio to Ref is shown in parentheses. Blank indicates that all generated codes have failed.}
\label{tab:perf_level3}
\begin{tabular}{l|l|r|rr|rr|rr|rr}
\hline
&&&\multicolumn{4}{c|}{NameToOptCcode}&\multicolumn{4}{c}{FrtcodeToOptCcode}\\\cline{4-11}
Routine&Parameters & Ref&\multicolumn{2}{c|}{GPT-4.1}&\multicolumn{2}{c|}{o4-mini}&\multicolumn{2}{c|}{GPT-4.1}&\multicolumn{2}{c}{o4-mini}	\\\hline
dgemm&transa=N, transb=N&2.7&17.4&(6.5x)&20.5&(7.7x)&18.0&(6.8x)&20.8&(7.8x)\\
&transa=N, transb=T&2.5&16.5&(6.5x)&&&16.1&(6.3x)&20.7&(8.1x)\\
&transa=T, transb=N&1.8&16.8&(9.3x)&0.8&(0.5x)&29.2&(16.1x)&23.8&(13.1x)\\
&transa=T, transb=T&0.3&15.7&(46.3x)&&&3.5&(10.4x)&20.3&(59.8x)\\
dsymm&side=L, uplo=L&3.6&13.6&(3.7x)&17.2&(4.7x)&19.9&(5.5x)&22.3&(6.2x)\\
&side=L, uplo=U&3.5&13.9&(4.0x)&17.3&(4.9x)&19.9&(5.7x)&22.4&(6.4x)\\
&side=R, uplo=L&2.7&13.4&(4.9x)&16.4&(6.0x)&19.0&(7.0x)&21.2&(7.8x)\\
&side=R, uplo=U&2.7&13.4&(4.9x)&21.3&(7.8x)&18.7&(6.8x)&21.1&(7.7x)\\
dsyrk&uplo=L, trans=N&1.0&&&2.9&(2.9x)&18.2&(18.2x)&16.6&(16.6x)\\
&uplo=L, trans=T&1.9&20.1&(10.6x)&21.8&(11.5x)&27.7&(14.7x)&23.0&(12.2x)\\
&uplo=U, trans=N&2.3&&&3.8&(1.6x)&16.5&(7.1x)&16.5&(7.1x)\\
&uplo=U, trans=T&1.9&36.5&(19.2x)&21.0&(11.0x)&34.7&(18.2x)&22.5&(11.9x)\\
dsyr2k&uplo=L, trans=N&1.8&&&&&31.0&(16.8x)&18.4&(10.0x)\\
&uplo=L, trans=T&3.1&42.9&(13.8x)&&&29.6&(9.5x)&23.1&(7.4x)\\
&uplo=U, trans=N&3.1&&&&&30.2&(9.9x)&18.1&(5.9x)\\
&uplo=U, trans=T&3.1&43.2&(13.7x)&&&27.4&(8.7x)&23.0&(7.3x)\\
dtrmm&side=L, uplo=L, trans=N, diag=N&3.1&&&1.6&(0.5x)&22.4&(7.4x)&&	\\
&side=L, uplo=L, trans=N, diag=U&3.1&&&1.3&(0.4x)&22.9&(7.4x)&23.0&(7.4x)\\
&side=L, uplo=L, trans=T, diag=N&1.8&&&&&28.7&(15.8x)&23.6&(13.0x)\\
&side=L, uplo=L, trans=T, diag=U&1.8&&&&&28.8&(15.7x)&23.7&(12.9x)\\
&side=L, uplo=U, trans=N, diag=N&3.0&&&1.7&(0.6x)&22.6&(7.5x)&22.4&(7.5x)\\
&side=L, uplo=U, trans=N, diag=U&3.0&&&1.7&(0.6x)&22.6&(7.6x)&22.2&(7.5x)\\
&side=L, uplo=U, trans=T, diag=N&1.1&&&&&27.7&(26.4x)&23.3&(22.2x)\\
&side=L, uplo=U, trans=T, diag=U&1.1&&&&&28.3&(26.7x)&23.6&(22.2x)\\
&side=R, uplo=L, trans=N, diag=N&3.0&&&&&0.3&(0.1x)&0.3&(0.1x)\\
&side=R, uplo=L, trans=N, diag=U&3.1&&&&&0.3&(0.1x)&0.3&(0.1x)\\
&side=R, uplo=L, trans=T, diag=N&3.4&&&&&0.3&(0.1x)&0.3&(0.1x)\\
&side=R, uplo=L, trans=T, diag=U&3.5&&&&&0.3&(0.1x)&0.3&(0.1x)\\
&side=R, uplo=U, trans=N, diag=N&3.3&&&&&0.3&(0.1x)&0.3&(0.1x)\\
&side=R, uplo=U, trans=N, diag=U&3.2&&&&&0.3&(0.1x)&0.3&(0.1x)\\
&side=R, uplo=U, trans=T, diag=N&3.3&&&&&0.3&(0.1x)&0.3&(0.1x)\\
&side=R, uplo=U, trans=T, diag=U&3.3&&&&&0.3&(0.1x)&0.3&(0.1x)\\
dtrsm&side=L, uplo=L, trans=N, diag=N&1.4&&&5.0&(3.5x)&20.8&(14.6x)&24.0&(16.9x)\\
&side=L, uplo=L, trans=N, diag=U&1.4&&&5.4&(3.7x)&20.9&(14.6x)&24.7&(17.2x)\\
&side=L, uplo=L, trans=T, diag=N&1.9&&&16.5&(8.8x)&23.6&(12.6x)&23.5&(12.6x)\\
&side=L, uplo=L, trans=T, diag=U&1.9&&&18.0&(9.6x)&23.6&(12.5x)&23.7&(12.6x)\\
&side=L, uplo=U, trans=N, diag=N&0.9&&&6.9&(8.0x)&20.7&(23.8x)&24.4&(28.0x)\\
&side=L, uplo=U, trans=N, diag=U&0.9&&&4.4&(5.0x)&20.9&(24.0x)&24.5&(28.1x)\\
&side=L, uplo=U, trans=T, diag=N&1.9&&&17.1&(9.2x)&23.4&(12.6x)&23.5&(12.7x)\\
&side=L, uplo=U, trans=T, diag=U&1.9&&&16.5&(8.9x)&23.4&(12.6x)&23.5&(12.7x)\\
&side=R, uplo=L, trans=N, diag=N&3.3&&&&&9.9&(3.0x)&0.3&(0.1x)\\
&side=R, uplo=L, trans=N, diag=U&3.2&&&&&9.1&(2.8x)&0.3&(0.1x)\\
&side=R, uplo=L, trans=T, diag=N&2.8&&&&&6.3&(2.3x)&20.5&(7.4x)\\
&side=R, uplo=L, trans=T, diag=U&2.8&&&&&5.8&(2.1x)&18.0&(6.5x)\\
&side=R, uplo=U, trans=N, diag=N&3.3&&&&&8.8&(2.7x)&0.3&(0.1x)\\
&side=R, uplo=U, trans=N, diag=U&3.3&&&&&9.9&(3.0x)&0.3&(0.1x)\\
&side=R, uplo=U, trans=T, diag=N&2.8&&&&&6.3&(2.3x)&20.2&(7.3x)\\
&side=R, uplo=U, trans=T, diag=U&2.8&&&&&6.5&(2.4x)&18.9&(6.8x)\\\hline
\end{tabular}
\end{table*}%%%%%%%%%%%%%%%%%%%%%%%%%%%%%%%%%%%%%%

%%%%%%%%%%%%%%%%%%%%%%%%%%%%%%%%%%%%%%%%%%%%%%%%%%%%
%%%%%%%%%%%%%%%%%%%%%%%%%%%%%%%%%%%%%%%%%%%%%%%%%%%%
%%%%%%%%%%%%%%%%%%%%%%%%%%%%%%%%%%%%%%%%%%%%%%%%%%%%
\section{Conclusion}
\label{sec:conclusion}
In this study, we evaluated the ability of OpenAI's GPT-4.1 and o4-mini models to generate BLAS routine codes. We observed that both LLM models were able to output correctly working C code from routine name alone in at maximum 10 trials for many BLAS routines. The generated code was not necessarily has the same structure of the reference Fortran code. This fact suggests that both models learned the routine specifications from documents and explanatory materials available on the Internet that contain the specifications of BLAS and used them to generate code.
\par %%%

This paper is a report of a case study but the case to generate BLAS code is not practical because open-source codes and highly optimized implementations of BLAS routines are available. However, the result shown in this paper suggests that the code generation by LLMs for other numerical codes with BLAS-level complexity is already in the practical stage if the specifications are written accurately. In addition, we have achieved a certain degree of code speed-up through basic optimizations including OpenMP parallelization, SIMD vectorization, and cache blocking, in some routines. It is interesting that even though the LLM models used in this study are generic and not specific to code generation, it produced the results shown in this report.  
\par %%%

There are various possible ways to improve the generated results. Prompts can provide more detailed information, or even without enhancing the LLM model, one can utilize RAGs to teach basic optimization methods, as in existing studies, or iteratively modify prompts or feed back bugs or obtained performance to improve the code. Along with the rapid performance improvement of generic models provided as a service, it is expected that these techniques can be used together to generate even higher-performance code.
\par %%%

% use section* for acknowledgment
\ifCLASSOPTIONcompsoc
  % The Computer Society usually uses the plural form
  \section*{Acknowledgments}
\else
  % regular IEEE prefers the singular form
  \section*{Acknowledgment}
\fi
This work used computational resources Flow Type II system provided by Information Technology Center, Nagoya University through Joint Usage/Research Center for Interdisciplinary Large-scale Information Infrastructures and High Performance Computing Infrastructure in Japan (Project ID: jh250015). Also, this work is supported by JSPS KAKENHI Grant Number JP23K11126 and JP24K02945.
\par %%%

\bibliographystyle{IEEEtran}
\bibliography{blasgen_mukunoki}

\end{document}